\documentclass{l4dc2025}

\usepackage{caption}
\usepackage{xcolor}

\title[Automating the loop in traffic incident management on highway]{Automating the loop in traffic incident management on highway}
\usepackage{times}

\author{%
 \Name{Matteo Cercola} \Email{matteo.cercola@polimi.it}\\
 \addr Dipartimento di Elettronica, Informazione e Bioingegneria, Politecnico di Milano, Milano, Italy.
  \AND
 \Name{Nicola Gatti} \Email{nicola.gatti@polimi.it}\\
 \addr Dipartimento di Elettronica, Informazione e Bioingegneria, Politecnico di Milano, Milano, Italy.
 \AND
  \Name{Pedro Huertas Leyva} \Email{pedro.huertasleyva@movyon.com}\\
 \addr MOVYON SpA (Gruppo Autostrade per l'Italia)
  \AND
  \Name{Benedetto Carambia} \Email{benedetto.carambia@movyon.com}\\
 \addr MOVYON SpA (Gruppo Autostrade per l'Italia)
 \AND
 \Name{Simone Formentin} \Email{simone.formentin@polimi.it}\\
 \addr Dipartimento di Elettronica, Informazione e Bioingegneria, Politecnico di Milano, Milano, Italy.
}

\begin{document}

\maketitle

\begin{abstract}%
 Effective traffic incident management is essential for ensuring safety, minimizing congestion, and reducing response times in emergency situations. Traditional highway incident management relies heavily on radio room operators, who must make rapid, informed decisions in high-stakes environments. This paper proposes an innovative solution to support and enhance these decisions by integrating Large Language Models (LLMs) into a decision-support system for traffic incident management. We introduce two approaches: (1) an LLM + Optimization hybrid that leverages both the flexibility of natural language interaction and the robustness of optimization techniques, and (2) a Full LLM approach that autonomously generates decisions using only LLM capabilities. We tested our solutions using historical event data from Autostrade per l’Italia. Experimental results indicate that while both approaches show promise, the LLM + Optimization solution demonstrates superior reliability, making it particularly suited to critical applications where consistency and accuracy are paramount. This research highlights the potential for LLMs to transform highway incident management by enabling accessible, data-driven decision-making support.
\end{abstract}


\section{Introduction}
Large Language Models (LLMs) are trained on vast datasets of text, learning to predict the next word in a sequence, which enables them to generate coherent and contextually relevant text. Their power lies in their ability to understand and generate human-like language, handle a wide range of tasks, and adapt to various contexts. LLMs excel at tasks that require natural language processing, including answering questions, summarizing information, and even performing basic reasoning by leveraging patterns learned during training.
Recent studies (\cite{wei2022chain,rao2023evaluating,sha2023languagempclargelanguagemodels,bian2024chatgptknowledgeableinexperiencedsolver,JMLR:v25:23-0870,fu2024drive}) have demonstrated that LLMs exhibit impressive reasoning capabilities. While their training is primarily based on language data, these models have shown the ability to make inferences, connect facts, and simulate logical reasoning to some extent. Recent studies demonstrate the potential of large language models to tackle complex language-based tasks such as planning (\cite{liu2023llmpempoweringlargelanguage,ahn2022icanisay,singh2023progprompt}), generating code (\cite{gur2024realworldwebagentplanninglong,10.1145/3511861.3511863,li2022competition}), and optimization (\cite{pryzant2023automaticpromptoptimizationgradient,10611913,nie2024importancedirectionalfeedbackllmbased,guo2024optimizinglargelanguagemodels,yao2024retroformerretrospectivelargelanguage}). Building on this progress, our research investigates the application of LLMs for decision-making in critical scenarios. Specifically, we propose two alternative approaches that integrate natural language interaction into decision-support systems:
\begin{itemize}
    \item LLM + Optimization: This hybrid approach combines the flexibility of LLMs with the robustness of classical optimization techniques, ensuring solution reliability while enabling a natural language interface.
    \item Full LLM: In this approach, decisions are generated exclusively by LLMs, offering a fully language-based solution without relying on external algorithms.
\end{itemize}
We have tested our proposed solutions on the problem of managing highway events, where decisions must be both efficient and reliable. By employing GenerativeAI these models can enable users to engage with the systems through
simple, conversational exchanges, thereby facilitating the interaction. Some works on this topic are \cite{grigorev2024incidentresponsegptgeneratingtrafficincident} and \cite{goecks2023disasterresponsegptlargelanguagemodels}, where they use LLMs to generate response plans based on accident descriptions by learning from a knowledge base of incident response guidelines.
A key advantage of our approach is the integration of real historical data, which includes detailed event descriptions and management records provided by Autostrade per l'Italia. This data-driven approach enables our systems to learn directly from historical patterns and responses, fostering decision-making grounded in actual operational experience. 
We evaluated both the Full LLM and LLM + Optimization solutions in the context of traffic incident management, and both showed promising results. However, given the critical nature of this application, we also assessed the reliability of each approach. Our findings indicate that the LLM + Optimization solution, by leveraging optimization techniques, offers more reliable outcomes, making it better suited for high-stakes environments where consistency and accuracy are crucial.
\section{Problem state and case description}
Effective management of highway events, such as traffic
accidents, breakdowns, or risky weather conditions, is crucial
to maintain traveler safety, minimize congestion, and ensure
smooth traffic flow. The ability to quickly and accurately
respond to these incidents can significantly speed up the
process of assisting injured people and reducing the risk of
secondary accidents. Currently, this critical responsibility
falls on the shoulders of radio room operators, who are tasked
with managing highway events in real-time.
Radio room operators are trained with lengthy manual procedures that are often not fully applicable in real-world scenarios.
The nature of their work demands high levels of attention
and quick decision-making, especially during long shifts that
are mentally and physically taxing. These operators must
assess a wide range of variables, from traffic patterns to
weather conditions, all while coordinating responses across
multiple stakeholders, including emergency services, road
maintenance teams, and police forces. Furthermore, the decisions they make are influenced by subjective factors, such as
their individual experience and interpretation of unfolding
events, which can lead to inconsistencies in how similar
incidents are handled. 

\noindent Innovative solutions are urgently needed to support operators in making rapid, consistent, and well-informed decisions, particularly in high-stakes environments where errors can lead to severe consequences. Given the critical and overwhelming job of the operators, such solutions should facilitate decision-making through natural and intuitive interaction, enhancing reliability without overburdening
the user.

\section{Methodologies}
The following are the methodologies we developed to tackle the problem:
\begin{itemize}
    \item \textbf{LLM + Optimization Block}: Combines the flexibility of large language models with optimization algorithms, ensuring that decisions are both contextually sound and operationally robust.
    \item \textbf{Full LLM}: Relies exclusively on LLMs to make decisions, for the purpose of testing them in decision problems.
\end{itemize}
Both methodologies take a natural language event description as input (e.g., type of event, number of vehicles involved, location, time, etc.) and produce a natural language output detailing the sequence of decisions to take (e.g., calling the police, closing a lane, etc.).
\subsection{LLM + Optimization Block}
\begin{figure}
    \centering
    \includegraphics[width=0.75\textwidth]{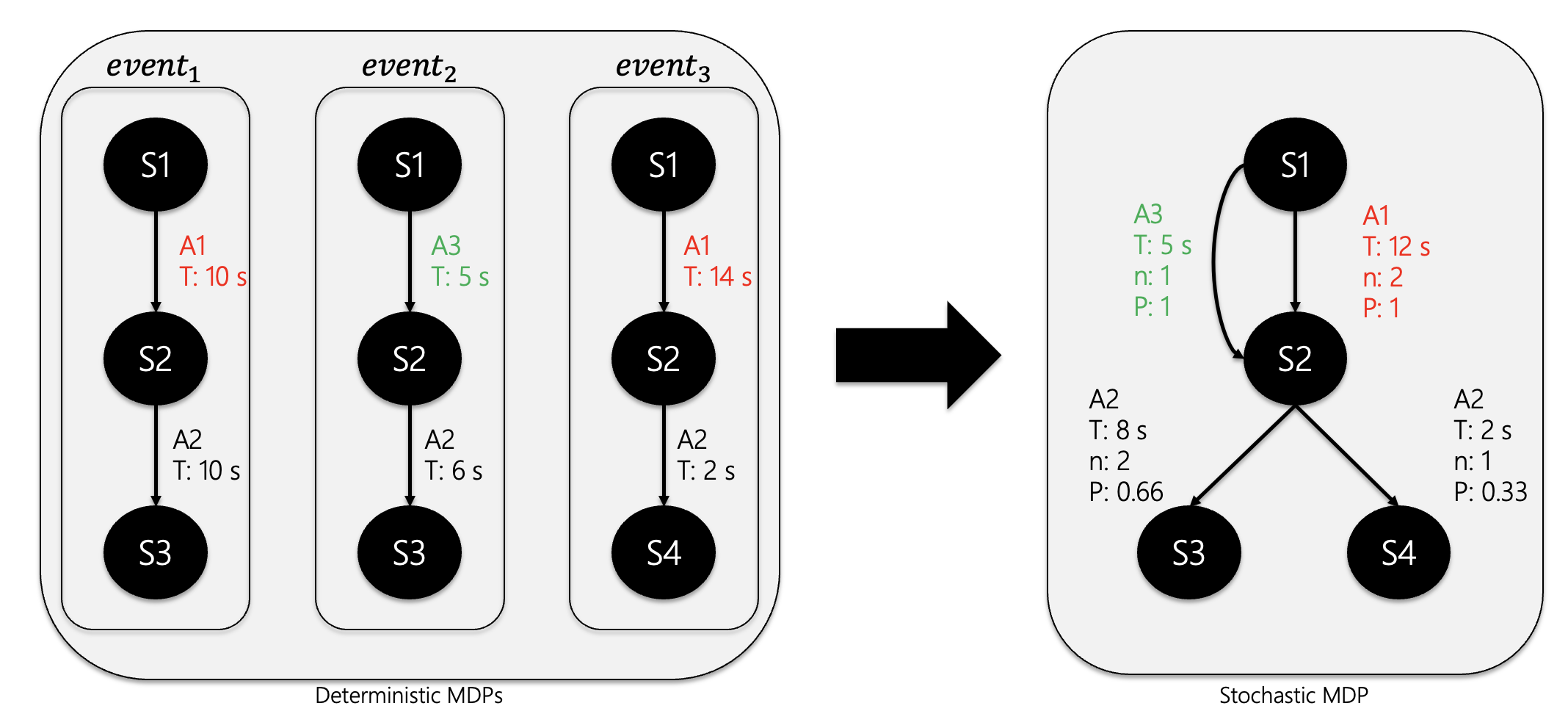}
    \caption{Graph representation of the system as a stochastic MDP, where nodes represent states and edges represent actions. Each edge is labeled with $n$, the number of times the action has been executed; $T$, the time associated with each action; and $P$, the probability of the action occurring.}
    \label{fig:Graph_representation}
\end{figure}

\begin{figure}[ht]
\includegraphics[width=0.6\textwidth]{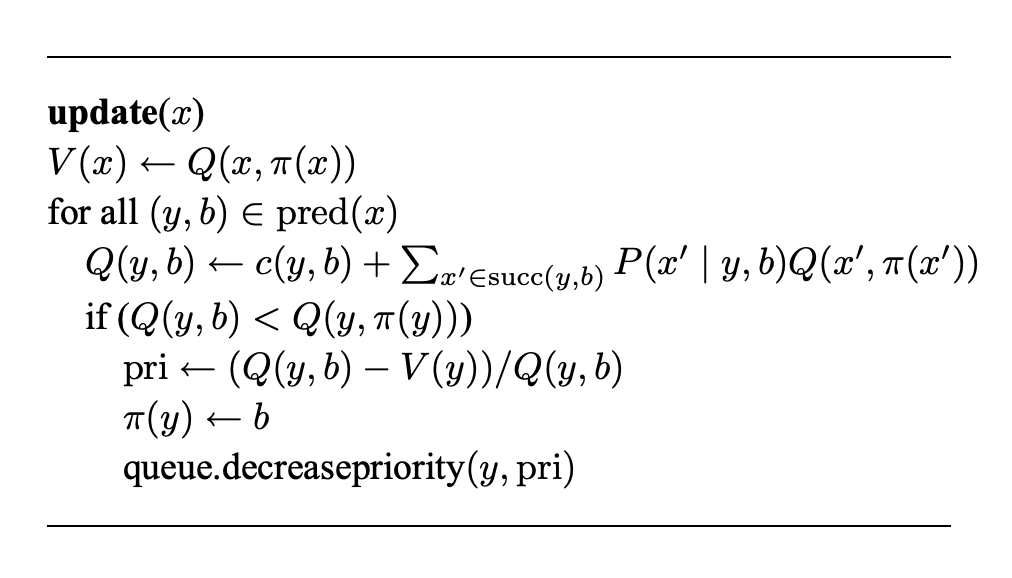}
    \caption{The update function for the Improved Prioritized Sweeping algorithm}
    \label{fig:IPS}
\end{figure}

\begin{center}
\noindent\begin{minipage}[t]{\linewidth}
    \centering
    \includegraphics[width=0.85\textwidth]{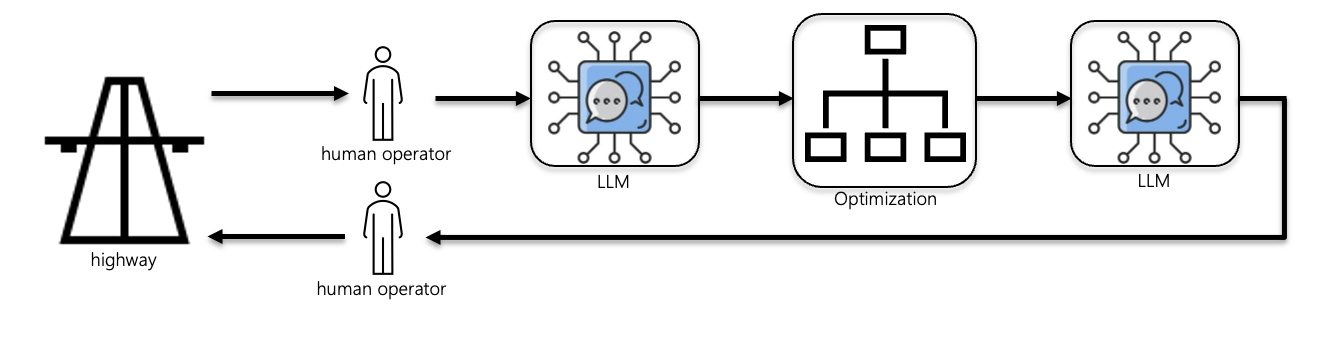}
    \captionof{figure}{LLM + Optimization Block pipeline.}
    \label{fig:LLM+Opt pipeline}
\end{minipage}
\end{center}
The proposed LLM + decision block solution combines the strengths of LLMs and optimization algorithms to establish a robust framework for managing highway events, building upon the approach presented in \cite{liu2023llmpempoweringlargelanguage}. LLMs excel at processing unstructured inputs and facilitating natural language interactions, while optimization algorithms provide mathematically grounded decision-making guarantees. By merging these two approaches, we aim to develop a system that is both user-friendly and theoretically reliable. The solution, visible in Figure \ref{fig:LLM+Opt pipeline}, includes two translation blocks (LLMs) at the beginning and end of the pipeline and a decision block in between. This structure enables radio room operators to engage with the optimization algorithm through natural, conversational interactions, thus reducing the technical complexity of the interface.\\
In this system, the two LLM components act as translators: the first converts natural language inputs into a structured format that can be processed by the optimization algorithm and the second retranslates the algorithm’s output back into natural language. This process effectively treats the input format of the optimization algorithm as a separate language, making the task of reformatting the highway event description a translation task, which LLMs are well-suited to handle. To achieve this, we utilized in-context learning (\cite{min2022rethinkingroledemonstrationsmakes}) and one-shot prompting. In-context learning allowed us to instruct the LLMs on the translation task without requiring any retraining, while defining the desired output format. Additionally, one-shot prompting (\cite{NEURIPS2020_1457c0d6}) was applied to provide the LLM with a single input-output example, demonstrating a resolution pattern. This technique has been shown to improve LLM performance by leveraging a simple example to guide the model's responses.\\
The optimization block handles the core decision-making process by learning from historical event management data provided by Autostrade per l'Italia, comprising thousands of real-world reports detailing highway events and their corresponding management.\\
Each report is modeled as a deterministic Markov Decision Process (MDP), where the states represent event descriptions (e.g., type of event, number of vehicles involved, location, time, etc...), and the actions are those taken to transition from one state to another (e.g., calling the police, closing a lane, etc...). The cost of each action is represented by the time it requires. Then, we create a single stochastic MDP by joining the nodes corresponding to the same event description, as shown in Figure \ref{fig:Graph_representation}. In the stochastic MDP representation, each transition has an associated probability and the cost function is updated by factoring in the frequency of past actions. The cost function prioritizes actions that have been executed more frequently and is shown in Equation \ref{eq:cost}: 
\begin{equation}\label{eq:cost}
c(s, a) = T(s, a) + \frac{penalty}{n(s, a)}    
\end{equation}
 where ${n(s, a)}$ is the number of times action $a$ has been executed in the state $s$ and $T(s,a)$ is the time action $a$ requires in state $s$. As a penalty, we choose the average time an action requires.
 \noindent
In formulating the problem as an MDP, determining the optimal action for each state corresponds to computing the optimal Q-value, $Q^*(s,a)$, for each state-action pair. This value represents the expected cost of reaching the goal from an initial state $s$, taking action $a$, and then following the optimal policy.  Thus, for any given state, the optimal action is identified as the one associated with the lowest Q-value.
To efficiently compute the optimal action for each state, we used the Improved Prioritized Sweeping (IPS) algorithm (\cite{mcmahan2005fast}). This algorithm extends to stochastic MDPs the priority queue idea of Dijkstra's algorithm, so keep states on a priority queue sorted by how urgent it is to expand them. Dijkstra's algorithm is guaranteed to find an optimal ordering for a deterministic positive-cost MDP, ensuring fast convergence of $Q$ to the optimal $Q^*$.\\
IPS does a backward search, so initially, the priority queue is populated with the end goal states, where the accident has been resolved. Iteratively the element with the lowest priority in the queue is visited by following the procedure outlined in Figure \ref{fig:IPS}.
\noindent
For each state $y$ that precedes $x$, the Q-value $Q(y,b)$ is computed. If the state $y$ is not already closed (\ref{eq:closed}), a priority value $pri$ is computed. 
\begin{equation}\label{eq:closed}
    Q(y,b) < Q(y,\pi(y))
\end{equation}
If $y$ is not yet in the queue, it is added with priority $pri$ otherwise its priority is just updated.
The priority function $pri$ has at the numerator the value change $\Delta \psi$ (\ref{eq:Delta}), which reflects the potential change in the state's value upon expansion and is non-positive. 
\begin{equation}\label{eq:Delta}
    \Delta \psi =Q(y,b) - V(y)
\end{equation}
$V(y)$ is the state value, which is the expected cost to reach the goal starting at state $y$. The priority function $pri$ is influenced by the the upper bound on priority ($P_{max}$) for the state-action pair (\ref{eq:UpperBound}), which is positive. 
\begin{equation}\label{eq:UpperBound}
P_{max} =\frac{1}{Q(y,b)}
\end{equation}
Thus, the overall priority function $pri$ is negative or zero and it decreases when the value change $ \Delta \psi$ increases and it increases as the upper bound $P_{max}$ increases, ensuring that high-priority states are processed promptly.
After executing IPS we know for each state in the MDP its optimal action. During inference, when presented with a new state description, we predict the optimal action by identifying the most similar node in the graph based on state features. Once the closest matching node is found, the associated optimal action can be directly inferred from it. To generate the full sequence of optimal actions, the process involves identifying the optimal action path leading to an end state (event resolved). This sequence is then refined by the final translational LLM, which converts it into a more human-readable and natural language response.
Selecting the most similar node in the MDP is crucial to ensure accurate action recommendations. We measure node similarity using a weighted distance metric, as formalized in Equation \ref{eq:Similarity}.
This approach accounts for the varying importance of features in the decision-making process. For instance, when deciding whether to call an ambulance, the presence of injured individuals carries more weight than the exact location in terms of kilometers. To assign feature weights effectively, we apply the Term Frequency-Inverse Document Frequency (TF-IDF) method. Term Frequency (TF) enhances the significance of features that occur frequently within a set of similar incidents, while Inverse Document Frequency (IDF) reduces the importance of features that are prevalent across multiple distinct incident types. For this analysis, we define a document as a set of reports that share the same resolution, that is, they show the same sequence of actions to handle the event. For each feature, the weight is the maximum TF-IDF value calculated across these reports, ensuring that the weight highlights features most relevant to the decision-making process.
The distance between two states, is then computed as follows:
\begin{equation}\label{eq:Similarity}
d(s,s^{\prime}) = \sum_{i=1}^{N}   w_i |f_{s,i} - f_{s^{\prime},i}| 
\end{equation}
where $f_{s}$ and $f_{s^{\prime}}$ are the feature vectors representing the states, with each element representing either a numerical feature value or a binary indicator (1 or 0) for categorical features. The weight vector 
$w$ is applied element-wise, reflecting the relevance of each feature in the distance calculation. At inference time, given a new state description, the optimal actions are determined by identifying the node with the smallest distance, calculated using Equation \ref{eq:Similarity}. 
Finally, to provide more information to the operator at decision time, two statistical forecasts are computed for each state: the predicted resolution time and the probability of a subsequent event occurring. These indices are crucial for the operator, as they offer approximated values that help inform decision-making during incident resolution. For example, knowing that an event might take hours to resolve might prompt an operator to take different actions than if it is expected to resolve in a few minutes. The same principle applies to event probability: for instance, understanding that an event is likely to generate heavy traffic allows the operator to take proactive measures to prevent congestion.
Both forecasts are computed using Improved Prioritized Sweeping, but with different cost functions. The time forecast uses the time each action requires as its cost function, while the prediction of future probable events relies on transition probabilities.
\subsection{Full LLM as a decision maker}
\begin{figure}[ht]
    \centering
    \includegraphics[width=0.65\textwidth]{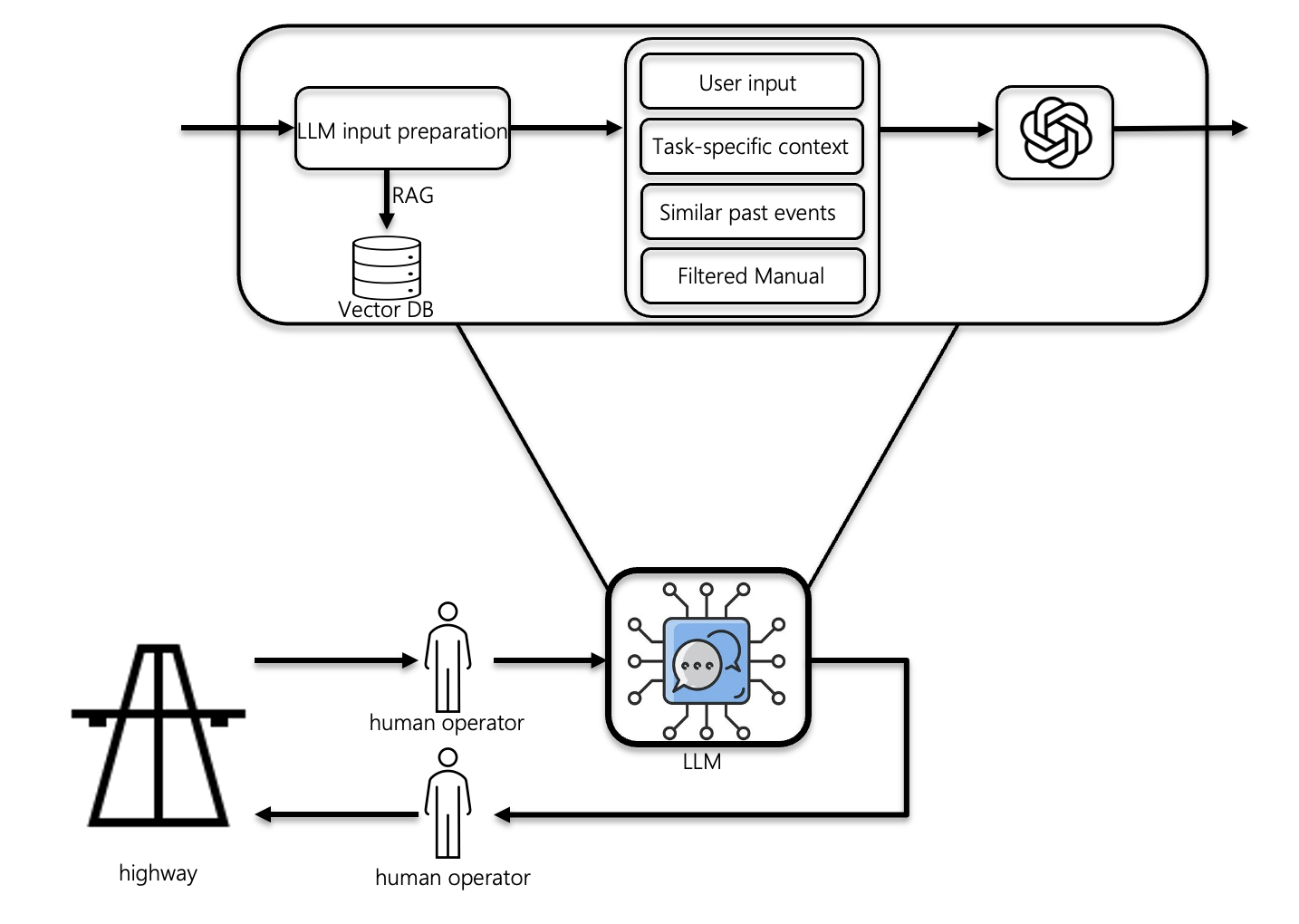}
    \caption{Full LLM pipeline.}
    \label{fig:Full LLM pipeline}
\end{figure}

We propose a solution where an LLM, GPT-4o mini, completely solves a decision-making problem for traffic incident management. The idea is to place the LLM in the role of a radio room operator, tasked with managing traffic incidents. To achieve this, we simulate the decision-making environment of the operator by providing the LLM with access to the same tools and information they use, such as the manual with procedures for traffic event management on highways and the historical event reports with information on the event's characteristics, the processes followed, and the time required.
To enhance its decision-making capabilities, we integrate Retrieval-Augmented Generation (RAG) (\cite{lewis2020retrieval}) into the solution. RAG allows the LLM to query and retrieve relevant information from an external database. We exploit RAG to let the LLM consult historical event reports and the manual with procedures.
Accessing a database of past traffic incident management cases enables the LLM to retrieve similar incidents from historical data to support decision-making for new events. 
The dataset of past events is split into chunks, with each chunk representing a past event description, its management, and the resolution time. 
Each chunk is stored in a vector database, where the event description is embedded using OpenAI's text embedding model, which serves as the basis for retrieval. The additional information, such as the event management and duration, is stored as metadata to maintain correspondence with the event description but is not used in the retrieval process. When a new event is presented to the Full LLM solution, the most similar events are retrieved based on cosine similarity applied to the embedded event descriptions.
Cosine similarity measures the cosine of the angle between two vectors, determining how closely aligned they are, with values ranging from -1 to 1. These similar past events, along with their corresponding management actions and durations, are then incorporated into the context provided to the LLM to support decision-making. To ensure that only relevant past events are provided to the LLM, we set a retrieval threshold based on an experimental analysis that balances the quality of retrieved information with the input prompt's length. Specifically, the threshold was chosen to include only events with sufficiently high similarity scores, filtering out less relevant cases while maintaining a manageable prompt size for effective LLM processing.
Providing information about the resolution of similar past events enables the LLM to leverage established strategies from prior cases, allowing it to understand how comparable situations were successfully managed and apply these insights to current scenarios. This approach mirrors the experience-based decision-making of human operators, who often rely on their knowledge and memory of how similar incidents were successfully managed.
To further enhance decision-making, we employed RAG to allow the LLM to consult the manual with procedures provided by Autostrade per l'Italia, a comprehensive 600-page document detailing procedures for various major events. Due to the manual’s length, incorporating the entire text into the LLM’s prompt would be impractical. RAG addresses this by enabling selective retrieval, providing only the most relevant sections of the manual as input to the LLM. The way the data is split and stored has a crucial effect on the quality of both retrieval and generation tasks. First, the manual was segmented based on its index. To further optimize each section, we applied semantic chunking (\cite{llamaindex_semantic_chunking}), which divides the text into meaningfully complete and contextually coherent pieces. Semantic chunking analyzes the sequence of sentences to determine optimal breakpoints where the text should be divided into two chunks. A breakpoint is inserted when the semantic distance between two consecutive sentences exceeds a threshold, indicating a significant shift in topic. This method ensures that the resulting chunks maintain semantic coherence and are contextually meaningful for retrieval. At decision time, the LLM is provided with only the most relevant parts of the manual to inform its recommendations. These parts are selected based on cosine similarity, with a threshold applied to the retrieval score to filter for the most pertinent information.\\
To enable the LLM to simulate the role of a radio room operator, we employed in-context learning (ICL) (\cite{min2022rethinkingroledemonstrationsmakes}). This approach allows us to clearly specify the task and seamlessly integrate all relevant filtered information from past events and the manual with procedures into the prompt. ICL allows pre-trained LLMs to address new tasks without the need for costly fine-tuning. Instead of retraining the model, ICL enables us to provide the LLM with task-specific instructions and relevant data as part of the input. In our case, we package all available information, including the description of the highway event, relevant historical data, and relevant sections of the manual allowing the LLM to process and respond as if it were a human operator making real-time decisions. This enables the system to provide context-aware, informed responses based on both current input and past experiences. This approach reduces complexity while maintaining flexibility in adapting the LLM to the task of traffic incident management.
Furthermore, we employed Chain of Thought (CoT) prompting to guide the LLM in decomposing the original problem into a sequence of reasoning steps, leading to the final solution. This approach enhances the model's reasoning capabilities, as demonstrated in previous work by \cite{wei2022chain}. The entire architecture of the Full LLM is illustrated in Figure \ref{fig:Full LLM pipeline}.
\noindent
Similar to the LLM + Optimization solution, the Full LLM solution also provides the two statistical forecasts previously defined: the predicted resolution time and the probability of a subsequent event occurring. This is achieved by using ICL to instruct the LLM model to generate these statistics, considering the most similar events to the input event, which are retrieved through RAG. 
\section{Experimental Results}
\noindent\begin{figure}
    \centering
    \includegraphics[width=0.8\textwidth]{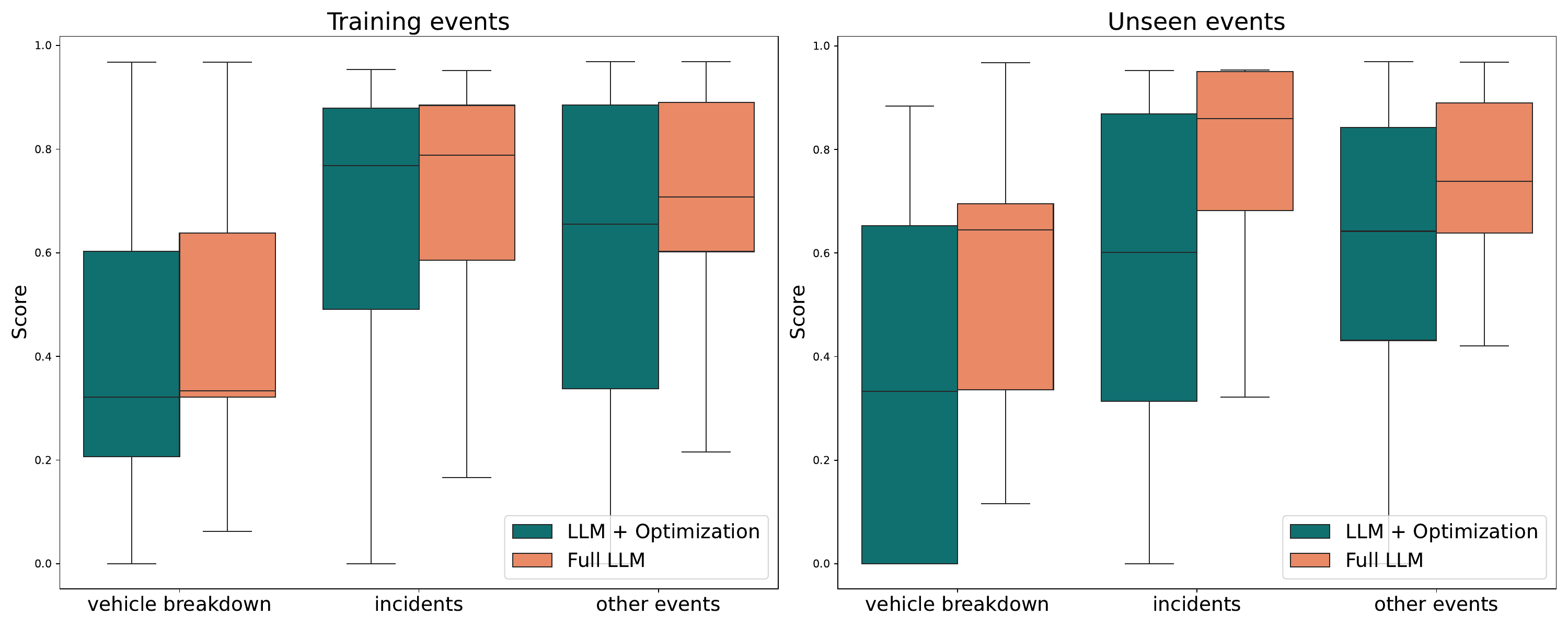}
    \caption{Box plot of evaluation scores for the two proposed solutions across three event types.}
    \label{fig:Evaluation}
\end{figure}
We tested our proposed solutions on the traffic incident management problem, where decisions need to be both efficient and reliable. The following metrics were used for performance assessment:
\begin{itemize}
    \item Score:  A numerical metric assessing the relevance of the decisions in relation to the procedures outlined in the manual.
    \item Consistency: A measure of the stability of predicted management actions under minimal perturbations to the input.
\end{itemize}
The dataset, provided by Autostrade per l'Italia, was partitioned into a training set (80\%) and a test set (20\%), to evaluate our solutions on both seen and unseen events during training. The training set was used to construct the MDP for the LLM + Optimization solution, while the Full LLM solution utilized this set through RAG. Testing was conducted across three primary categories of highway events: vehicle breakdowns and vehicle collisions, which are the events that happen most frequently, and a third category, other events, that contains other types of less frequent events, which are not necessarily less critical. Metrics are computed by inputting 50 events per category from the training set and 50 events per category from the test set (unseen events).
\subsection{Score metric}
We measured the accuracy of the predictions generated by the two algorithms against the manual of procedures. The evaluation score, shown in Equation (\ref{eq:Score}), quantifies the relevance of predicted actions:
\begin{equation}\label{eq:Score}
score(x) = \frac{1}{N} \sum_{i=1}^{N} \frac{\textit{cosine}\left(M_i, A_i\right) - \textit{cosine}\left(M_i, rnd_{M_i}\right)}{1 - \textit{cosine}\left(M_i, rnd_{M_i}\right)}  
\end{equation}
The score $score(x)$, as shown in Equation (\ref{eq:Score}), represents the average cosine similarity ($cosine$) between the embedded actions specified for event $x$ by the manual with procedures, denoted as $M$, and the embedded actions predicted by our algorithms for $x$ denoted as $A$. This score is normalized by a baseline similarity, calculated as the cosine similarity between $M_i$ and a randomly generated string $rnd_{M_i}$, where $rnd_{M_i}$ is sampled from a uniform distribution over uppercase and lowercase letters and has the same length as the embedding of $M_i$.
To avoid accounting for the ordering of actions, the score is computed for every possible permutation of the predicted actions, with only the highest score being retained. The parameter  $N$ represents the number of actions specified in the manual, and it serves also to penalize events where the predicted number of actions differs from the expected count.
Figure \ref{fig:Evaluation} illustrates the evaluation results. Both solutions perform well when evaluated against the manual with procedures. Notably, lower performance is observed only in vehicle breakdown events, which requires some context. According to the manual, a general vehicle breakdown involves several actions to resolve the incident. However, in practice, some breakdowns resolve without operator intervention, such as when a stopped vehicle is able to restart autonomously. Consequently, in our experiments, situations where the model predicts 'take no action' receive a lower score, which frequently occurs with vehicle breakdown events. 
An advantage of the Full LLM solution is its ability to leverage the manual with procedures, whereas the LLM + Optimization model relies solely on historical data. This distinction is crucial in the evaluation framework we developed. One consideration is that we did not explicitly specify whether the Full LLM solution should prioritize similar past cases or the manual’s procedures at decision time. Based on the results, it seems that the LLM tends to prioritize the manual guidance for never-before-seen events, while it draws more on past cases for events belonging to the training set, which is a satisfactory result.
\subsection{Consistency metric}
\begin{figure}[ht]
    \centering
    \includegraphics[width=0.7\textwidth]{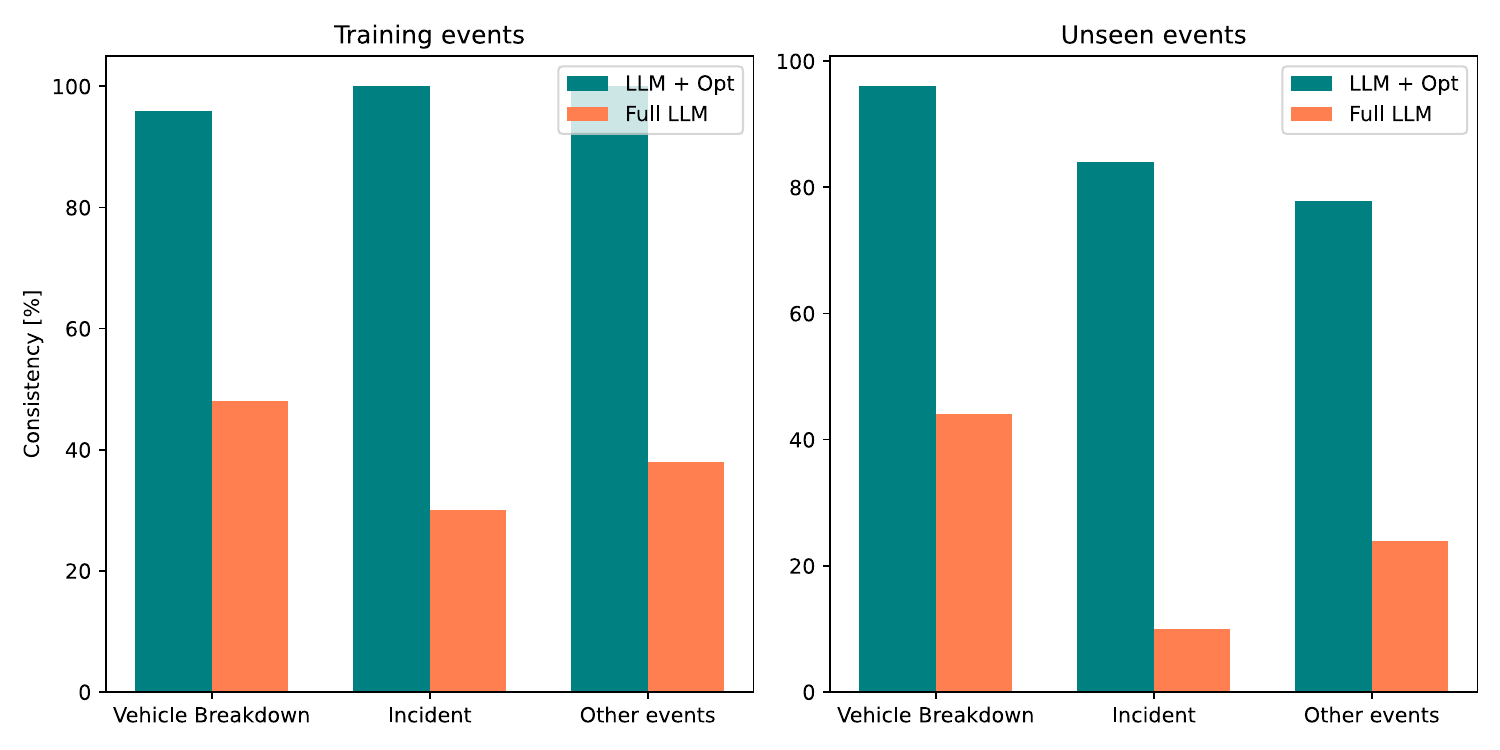}
    \caption{Comparison of the reliability of the Full LLM and LLM + Optimization solutions based on the Consistency [\%] metric. Higher percentages indicate greater reliability.}
    \label{fig:Consistency}
\end{figure}

While the Full LLM solution shows promise in its ability to assist or even automate decision-making in traffic incident management, there are important considerations regarding its reliability. One major limitation is that LLMs lack theoretical guarantees on the correctness of their output, so the decisions proposed by the Full LLM solution are inherently unpredictable and can vary based on subtle differences in input context, as shown in Figure \ref{fig:Consistency}. To evaluate this behavior, we employed the Consistency [\%] metric, which quantifies the stability of predicted management actions in response to minimal perturbations in the input. Specifically, these perturbations involve modifying certain event features to alter the state description while preserving the expected management action.
This unpredictability of the Full LLM solution raises concerns about deploying the solution in high-stakes environments where consistency and correctness are critical like in the highway management scenario. While the Full LLM approach offers substantial potential, the hybrid solution, LLM + Optimization, stands out as the more promising option due to its flexibility and performance reliability. 
\noindent
\section{Final discussion}
Our research demonstrates the feasibility and effectiveness of integrating Generative AI into decision-support systems, particularly for traffic incident management. The LLM + Optimization approach, which combines language processing capabilities with robust decision-making frameworks, offers a practical solution that enhances decision accuracy while simplifying user interaction. In contrast, while the Full LLM model performed well, its lack of robust guarantees underscores the need for refinement, especially in high-stakes scenarios where errors must be minimized. \\
Future research can focus on addressing the limitations of both models to enhance their effectiveness in critical traffic incident management scenarios. For the Full LLM model, improving reliability is a key goal. A possible approach, although complex in highway management, is to introduce a validation phase to verify decisions before finalization, similar to testing in code generation. Another valuable enhancement could be to provide explainability for the LLM’s decisions. Highlighting the features or past events that significantly influenced the decision would allow operators to assess the reliability of the LLM’s recommendations. Additionally, fine-tuning the LLM with domain-specific data, rather than relying solely on prompt engineering, could yield a model better suited to decision-making tasks, improving both performance and reliability, although the cost of training would increase significantly. For the LLM + Optimization model, the primary challenge lies in managing graph complexity. Clustering techniques within the MDP can be used to group similar states, streamlining computation and enhancing generalization. To further refine decision-making, we are exploring similarity metrics. Specifically, we are developing a Siamese network to learn an effective distance metric between states and plan to investigate additional techniques to further optimize performance. \\
In conclusion, this study represents a significant step toward intelligent, reliable, and user-friendly decision-support systems, with the potential to greatly enhance operational efficiency and consistency in traffic incident management.

\bibliography{main.bib} 

\end{document}